\documentclass[conference]{IEEEtran}
\usepackage{graphicx}
\usepackage{amsmath,mathrsfs}
\usepackage{subfigure, cite, url}
\usepackage{floatrow}
\DeclareMathOperator*{\argmin}{arg\,min}
\newfloatcommand{capbtabbox}{table}[]

\ifCLASSINFOpdf
\else
\fi
\begin{document}
%
\title{StackInsights: Cognitive Learning for Hybrid \\Cloud Readiness}

\author{
\IEEEauthorblockN{Mu Qiao, Luis Bathen, Simon-Pierre G\'{e}not, Sunhwan Lee, Ramani Routray} 
\IEEEauthorblockA{IBM Almaden Research Center\\650 Harry Road\\San Jose, California, CA} 
\IEEEauthorblockA{
	Email: {\{mqiao,bathen,sgenot,shlee,routrayr\}@us.ibm.com}
	}
}


%


\maketitle

\begin{abstract}
Hybrid cloud is an integrated cloud computing environment utilizing a mix of public cloud, private cloud, and on-premise traditional IT infrastructures. Workload awareness, defined as a detailed full range understanding of each individual workload, is essential in implementing the hybrid cloud. While it is critical to perform an accurate analysis to determine which workloads are appropriate for on-premise deployment versus which workloads can be migrated to a cloud off-premise, the assessment is mainly performed by rule or policy based approaches. In this paper, we introduce StackInsights, a novel cognitive system to automatically analyze and predict the cloud readiness of workloads for an enterprise. Our system harnesses the critical metrics across the entire stack: 1) infrastructure metrics, 2) data relevance metrics, and 3) application taxonomy, to identify workloads that have characteristics of a) low sensitivity with respect to business security, criticality and compliance, and b) low response time requirements and access patterns. Since the capture of the data relevance metrics involves an intrusive and in-depth scanning of the content of storage objects, a machine learning model is applied to perform the business relevance classification by learning from the meta level metrics harnessed across stack. In contrast to traditional methods, StackInsights significantly reduces the total time for hybrid cloud readiness assessment by orders of magnitude. 
\end{abstract}


%
\IEEEpeerreviewmaketitle

\section{Introduction}

Hybrid cloud, which utilizes a mix of public cloud, private cloud, and on-premise, has become the dominant cloud deployment architecture for enterprises. Public cloud offers a multi-tenant environment, where physical resources, such as computing, storage and network devices, are shared and accessible over a public network, whereas private cloud is operated solely for a single organization with dedicated resources. Hybrid cloud inherits the advantages of these two cloud models and allows workloads to move between them according to the change of business needs and cost, therefore resulting in greater deployment flexibility. The global hybrid cloud market is estimated to grow from USD 33.28 Billion in 2016 to USD 91.74 Billion in 2021~\cite{mandm2016}. 

Business sensitivity is one of the main factors that enterprises consider when deciding which cloud model to deploy. For example, an enterprise can deploy public clouds for test and development workloads, where security and compliance are not an issue. However, it is hard to meet PCI (payment card industry) or SOX (Sarbanes-Oxley) compliance in public clouds due to the nature of multi-tenancy. On the other hand, because private clouds are dedicated to a single organization, the architecture can be designed to assure high level security and stringent compliance, such as HIPAA (health insurance portability and accountability act). Therefore, private clouds are usually deployed for business sensitive and critical workloads. Infrastructure is another important factor to consider when choosing between public and private clouds. Since private cloud is a single-tenant environment where resources can be specified and highly customized, it is ideal to host data which are frequently accessed and require fast response time. For example, high-end storage system can be used in private cloud to deliver IOPS (input/output operations per second) within a guaranteed response time. 

\begin{figure}[h]
	\centering
	\includegraphics[width=0.77\textwidth]{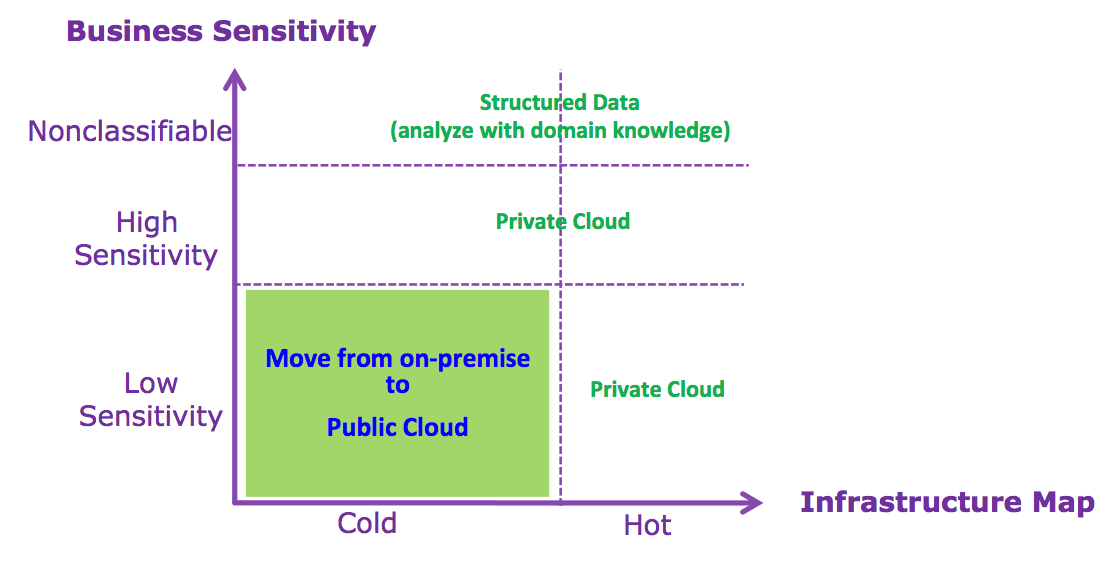}
	\caption{Hybrid cloud migration overview}
	\label{fig:intro}
\end{figure}

Moreover, business sensitivity and infrastructure are traditionally considered in two separate schools of work. However, not all data is created equal, neither is the infrastructure. In this paper, we introduce StackInsights, a novel cognitive learning system to automatically analyze and predict the cloud readiness of workloads for an enterprise by considering both business sensitivity and infrastructure. StackInsights classifies the entire data into several subspaces, as shown in Figure~\ref{fig:intro}, where the $X$-axis indicates the infrastructure heat map (e.g., storage access intensity) and the $Y$-axis represents the business sensitivity. A threshold on the $X$-axis is set to determine if the data is ``\emph{cold}" or ``\emph{hot}" with respect to infrastructure related performance metrics, and on the $Y$-axis, the data is classified into three categories: ``\emph{sensitive}", ``\emph{non-sensitive}", or ``\emph{non-classifiable}''. Formally, we define sensitive data as the data owned by the enterprise, which if lost or compromised, bares financial, integrity, and compliance damage. There are many different forms of sensitive data, such as sensitive personal information (SPI), personal health information (PHI), confidential business information, client data, intellectual property, and other domain-specific sensitive information. The category of ``\emph{non-classifiable}'' includes structured data, such as databases, the sensitivity of which can be analyzed using domain knowledge. For example, the databases storing employment information in the HR department should be highly sensitive. All the data which are cold and non-sensitive can be migrated to public clouds while the rest should reside in private clouds. The thresholds on the $X$-axis and $Y$-axis can also be adjusted by users. The areas of the subspaces indicate the size of data migrating to different clouds, thereby, serving as a cloud sizing tool. The hotness of data on the $X$-axis can be obtained by measuring infrastructure performance metrics. The key issue therefore lies in how to determine the business sensitivity of data on the $Y$-axis. 

To the best of our knowledge, StackInsights is the first cognitive system that uses machine learning to understand data sensitivity based on metadata, as it correlates application, data, and infrastructure metrics for hybrid cloud readiness assessment.  In contrast to traditional methods which require content scanning for sensitivity analysis, StackInsights significantly reduces the total running time through machine learning. It advises users what data are appropriate to be stored on premises, or to be migrated to the cloud, and the specific cloud deployment model, by integrating both data sensitivity and hotness in terms of infrastructure performance. 

The rest of the paper is organized as follows. We describe our motivation and contribution in Section~\ref{motivation}. The relevant work is reviewed in Section~\ref{related}. In Section~\ref{framework}, we introduce the framework of StackInsights as well as the cognitive learning components. Section~\ref{exp} is on the experiments and results. Finally, we conclude in Section~\ref{conclusion}.

\section{Motivation and Contribution}
\label{motivation}
 
The classification of data sensitivity belongs to the general domain of data classification, which allows organizations to categorize data by business relevance and sensitivity in order to maintain the confidentiality and integrity of their data. Data classification is a very costly activity. In large organizations, data is usually stored and secured by many repositories or systems in different geo locations, which may have different data privacy and regulatory compliances. Various security access approvals have to be obtained in order to get access to data. In addition, traditional sensitivity assessment approaches require an intrusive and in-depth content scanning of the objects, which is not scalable in this big data era, where numerous structured and unstructured data are generated in real-time. To solve this issue, we develop a machine learning model in StackInsights, which can perform a business sensitivity classification by learning from file metadata, which is much easier and cost-efficient to collect. By using meta data, we can already obtain a sensitivity prediction model with high accuracy. Therefore, we do not have to perform a detailed content analysis on all the files. Instead, intensive content analysis can be conducted on the predicted non-sensitive files for further screening if needed. Our model-based approach significantly reduces the total sensitivity assessment time. 

When migrating workloads among private, public, and hybrid clouds, one of the biggest challenges is the storage layer. An enterprise's infrastructure might consist of a mixture of file, block, and object storage, which have different properties and offer their own advantages.  Enterprises big or small tend to manage large and heterogeneous environments. For example, one of our IT environments supports around 100 business accounts, spread over several geo locations, amassing a total of 200 PB of block storages alone. Similarly, our file storage fabric is also massive, where file shares (mapped to volumes/q-trees) may be in the TB, or even PB scale. 

Given such a large storage infrastructure with a number of volumes, we need to determine their cloud migration priority, i.e., which volumes should be migrated first. Data sensitivity is one of the most important factors in cloud migration. Volumes of low sensitivity can be sanitized first and then migrated to the public cloud. From a cloud migration service admin point of view, it is not critical to know the exact sensitivity of the volumes, but rather the sensitivity ``level'' of the volumes, so that a priority can be given to a large number of volumes. Traditional sensitivity assessment approaches, which require content scanning, are very expensive. It is impractical and not necessary to perform a full content scanning on all the volumes in order to obtain the priority. Machine learning can help predict the sensitivity of files based on the easily collected file meta data, and then obtain the migration priority within a much shorter time. 

As it is expensive to determine the business sensitivity of each storage volume, we develop a clustering component in StackInsights, which identifies groups of volumes that share similar characteristics. Specifically, the volumes are clustered based on their meta level information, which are obtained by aggregating the file metadata at the volume level. The sensitivity of a representative volume in each cluster is used as the representative sensitivity of all the volumes in the same cluster. To obtain the sensitivity of a representative volume, we apply the previously introduced machine learning model to predict the sensitivity of each single file on that volume. The sensitivity of a volume is defined as the number of sensitive files divided by the total number the files. 

Similarly, we also obtain the IOPS of each volume and compute its IO density, which is defined as IOPS per GB. All the volumes with both low business sensitivity and IO density can be candidates to be migrated to public clouds while the other volumes should remain on premise or be migrated to private clouds.

\section{Related Work}
\label{related}
In the marketplace of enterprise softwares, there are tools developed for data classification in regard of data governance or life cycle management. For example,~\cite{symantec}~\cite{varonis} provide data classification services for managing and retaining various types of data such as emails and other unstructured data through pre-determined rules or dictionary searches. Data privacy and security have become the most pressing concerns for organizations. To embrace the newly announced General Data Protection Regulation (GDPR) by European Union, enterprises are making great efforts in addressing key data protection requirements as well as automating the compliance process. For example, IBM Security Guardium solutions~\cite{ibm} help clients secure their sensitive data across a full range of environments. Data classification, as the first step to security, has become extremely important. Only if we understand which data are sensitive through classification, we can design better security product to protect them. On the other hand,~\cite{gravitant} assesses the cloud migration readiness by providing a questionnaire to the owner of the infrastructure. Many of existing tools lack of cognitive aspects, and even if there is, the data preparation step requires the scanning of file content, which is not scalable or limits the type of files to which the tool can be applied.

Besides the rule-based approaches to classify data, there are previous attempts leveraging a predictive model. Model-based approaches are much more systematic and scalable because there is no need to generate a rule to classify files manually. Data classification based on the proof-of-concept system that crawls all files in order to analyze data sensitivity was studied in~\cite{park:2011}. A nearest neighbor algorithm was proposed in~\cite{Ali2015} to attribute the confidentiality of files. A general-purpose method to automatically detect sensitive information from textual documents was presented in~\cite{Sanchez2012} by adapting the information theory and a large corpus of documents. The application to data loss prevention by using automatic text classification algorithms for classifying enterprise documents as either sensitive or non-sensitive was introduced in~\cite{Hart2011}. However, most of these previous works require an exhaustive process to crawl the contents of data, which is impossible in many applications due to privacy, governance, or regulations. \cite{Mesnier2004} proposed to use the decision tree classifier for finding associations between a file's properties and metadata. 

There are preliminary works in the field of hybrid cloud migrations whose components include a data classification method. The tool for migrating enterprise services into hybrid cloud-based deployment was introduced in~\cite{Hajjat2010}. The complexity of enterprise applications is highlighted and the model, accommodating the complexity and evaluating the benefits of a hybrid cloud migration, is developed. Though the work shed insight on security of data, the tool does not assess the sensitivity of applications in the level of file. Rule-based decision support tool is deployed in~\cite{Khajeh2011} for the purpose of providing a modeling tool to compare various infrastructures for IT architects. In many practical cases, however, IT architects are blind on the contents of data and thus it is not straightforward to model their applications, data, and infrastructure requirements without understanding the nature of data such as sensitivity. In addition,~\cite{Menzel2012}  developed a framework which automates the migration of web server to the cloud.

As observed in previous works, data classification and hybrid cloud migration are explored separately although two components are tightly related. In contrast to the existing works, our proposed framework covers the whole process including classifying data, assessing the readiness of cloud migration, and finally a decision support for hybrid cloud migration. Furthermore, we develop an efficient and scalable method to determine cloud readiness by considering data sensitivity and infrastructure performance through a cognitive learning process.

\section{StackInsights Framework}
\label{framework}

\begin{figure*}[h]
\begin{center}
\includegraphics[width=6.0in]{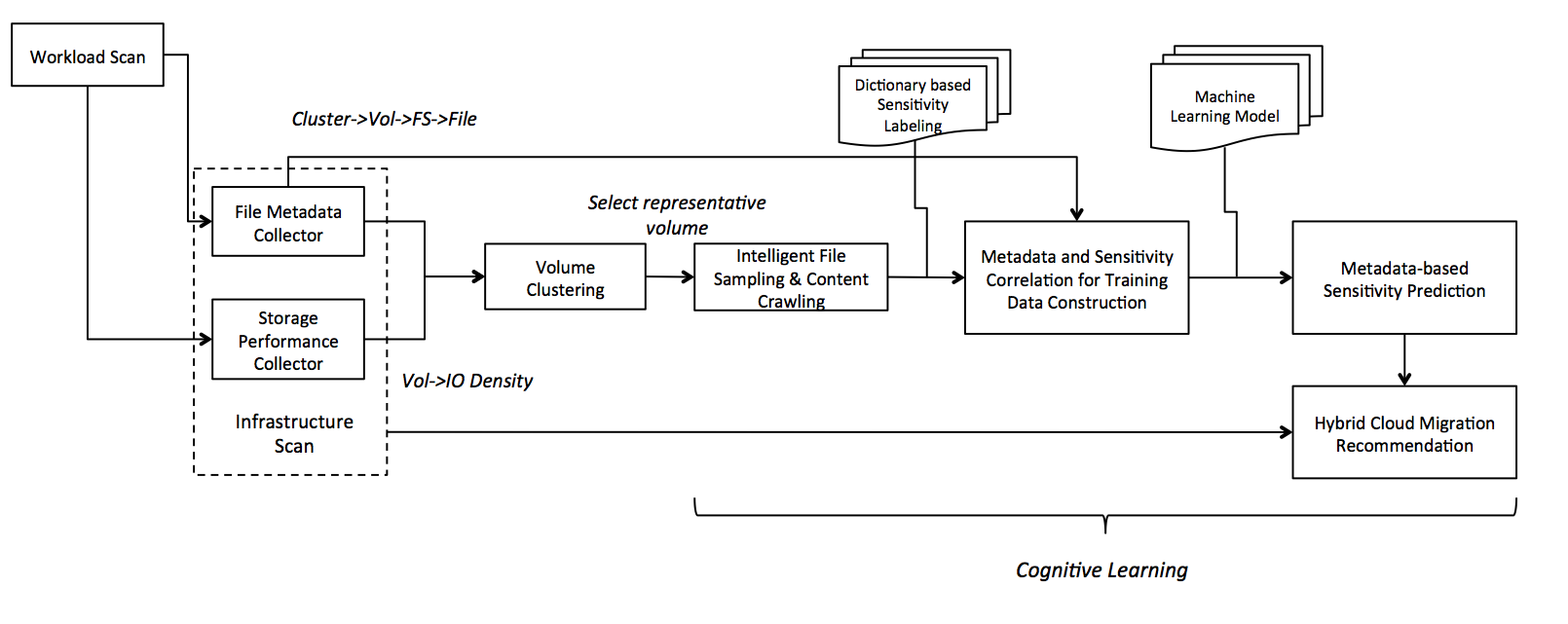}
\caption{StackInsights framework}
\label{fig:arch}
\end{center}
\end{figure*}

We show the high-level framework of StackInsights in Figure~\ref{fig:arch}. In order to gain insights into an existing IT environment, we scan various layers across the entire stack: 1) the application layer, 2) the data layer, and 3) the infrastructure layer. The application layer tells us the types of the running workloads, what components they depend on, and the specific requirements. The data layer provides file metadata as well as content. Finally, the infrastructure layer provides performance metrics, such as, how often the data is accessed, and where it is stored. 

\subsection{Workload Scan}

Before scanning the infrastructure and the data itself, we first need to understand what type of workloads are running within the enterprise. This can be done through the use of tools such as IBM's Tivoli Application Dependency Discovery Manager (TADDM) \cite{taddm}, which provides an automated discovery and application mappings. Additionally, the IT staff within each enterprise are also good sources as they are often the subject matter experts and can give a high-level overview of their workloads. This step is critical as we need to understand the workloads or applications before starting any sort of scan. For example, file content scanning could be invasive to latency-sensitive workloads and interfere with their running.

\subsection{Infrastructure Scan}
\label{scan}

In order to understand an IT environment, we need to get a picture of where the data is, how they can be accessed, and what types of storage the infrastructure consists of. Our framework is composed of a set pluggable modules that can be adapted to scan different infrastructures. This is important as infrastructures tend to be heterogeneous in nature. For example, a storage infrastructure may have a mixture of different storages, management technologies developed by multiple providers, as well as different ways of accessing data (e.g., via block, file, or object stores). For example, if the workload scan tells us that the storage layer consists mostly of storage filers, we can assume that at this layer, most data is reached through protocols such as Network File System (NFS), and Common Internet File System (CIFS), as well as manufacturer-specific APIs such as NetApp's ONTAP \cite{ontap}. Once a scan of the infrastructure has been completed, we can then build a location map of all the data. This step is critical as we need to be able to identify the storage volumes or data shares that are most critical for scanning. Similarly, we need to identify volumes/file shares that may host critical data or data administrators do not wish to migrate. For the rest of the paper, we will use $volume$ and $file~share$ interchangeably as NetApp filers have the notion of Q-trees/Volumes as mapping points for file shares (which are exposed to users through protocols such as CIFS), the root volume/q-tree for each share is mounted on our virtual machines as read-only directories. Similarly, with block storage, we care mostly about volume granularity, as most migration utilities operate at this level. 

\subsection{Volume Clustering}
Given a large heterogeneous IT infrastructure, we first apply a clustering method to identify groups of volumes that share similar characteristics. The volume are clustered based on their meta level information which are obtained by aggregating the metadata of all the files on the same volume. Each volume is represented by a feature vector. Note that more meta-level information about the volumes can be collected and included as additional features. We apply the K-means algorithm to obtain the volume clusters. Given a set of data points $(x_1, x_2, ..., x_n)$, where each data point is represented by a $d$-dimensional feature vector, K-means aims to partition the $n$ data points into $k$ sets: $S =  (s_1, s_2, ...,  s_k), k \leq n $, so that the total within cluster distance is minimized, i.e., its objective function is $\argmin_{S} \sum_{i=1}^{k}\sum_{x \in s_i} || x - \mu_{i} || ^2$, where $\mu_i$ is the mean of data points in $s_i$. 

After all the volumes are clustered, we select a representative volume from each cluster, which is defined as the one with minimum total distance to all the other volumes in the same cluster. We further analyze the business sensitivity of every representative volume. We assume that volumes in the same cluster share a similar sensitivity score. 

\subsection{Cognitive Learning}
Business sensitivity is a critical factor in hybrid cloud migration. We need to analyze the sensitivity of the representative volume of each volume cluster,  which is defined as the number of sensitive files divided by the total number of files on that volume. The current approach to detecting sensitive files requires in-depth content scanning, which is intrusive and expensive. However, even a single storage volume may contain millions of files, in TB or even PB scale. It will take a tremendous amount of time to do a full content scan in order to determine the sensitivity of all the files. In StackInsights, we develop a cognitive learning component to predict the sensitivity of files based on easily obtained metadata, which significantly reduces the total running time.

\subsubsection{File Content Crawling}
\label{datamap}
We randomly sample a subset of files from the selected representative storage volume, crawl their content, and apply the traditional rule based approach to determine the sensitivity. Files are identified as sensitive or non-sensitive by matching against a list of regular expressions and keywords predefined by users. Table~\ref{tb:sendic} shows a sample dictionary of sensitive information. 

\begin{table}
        {\small
	\begin{center}
	  \caption{Dictionary of sensitive information}   
	  \label{tb:sendic}
	  \begin{tabular}{| c | c |}
	  \hline
	   Email address & regex  \\ \hline
	   Phone number & regex  \\ \hline
	   Social Security Number & regex  \\ \hline
	   Credit card number  & regex \\ \hline
	   Keywords & list of tokens \\ \hline 
	  \end{tabular}
	\end{center}
	}
\end{table}

The definition of sensitive files shown above can be modified or extended with the domain knowledge of specific industry verticals. For instance, financial, healthcare, or retail industries may have their own guidelines to define sensitive files but we tried to come up with a reasonable criteria to single out sensitive files. The crawling output contains attributes such as file name, file path, flag for whether the content was crawled or not, the number of total tokens excluding stop words, the number of matching key words, email address, phone number, social security number, and credit card number. Users can define the file sensitivity labeling rule. For example, a file can be labelled as sensitive if it contains any sensitive information in the dictionary. Users can also specify a more stringent rule, such as the file is sensitive only if the percentage of sensitive information is above a certain threshold. After the content crawling process, each sampled file is labeled as one of the three classes, sensitive, non-sensitive, or unknown (in case they cannot be scanned, for example, unsupported file formats and encrypted files). The sensitivity labels of these files are correlated with their metadata, which compose the training data for building our sensitivity prediction model. 

\subsubsection{Intelligent File Sampling}
\label{sampling}
After the training data is prepared, we can build a binary classification model and apply it to predict the sensitivity of remaining files on the volume based on their metadata. However, it is well know that the quality of training data has a significant impact on the performance of machine learning model. In Section~\ref{datamap}, to prepare the training data, we randomly sample a subset of files, crawl their content, and determine the sensitivity of these files. One question remains: how should we conduct the random sampling in order to obtain a ``good'' training data? In StackInsights, we develop a clustering based progressive sampling method to solve this problem. 

All the files on a storage volume are first clustered using their metadata, for example, via K-means. We then compute the percentage of data points assigned to each cluster. A random sampling is performed on each cluster to select data points proportionally, with respect to the previously obtained percentages. For example, suppose the total number of files on the volume is 160,000, the number of files in a particular cluster takes 20\%, and we want to randomly sampling 3\% of the entire data as training. The final number of sampled files from that cluster is therefore $160,000 \times 20\% \times 3\% = 960$. We crawl the content of the selected files and determine their sensitivity using the approach introduced in Section~\ref{datamap}. 

A progressive sampling method is applied to determine the final total sampling size. We start from a relatively small sampling percentage and apply the aforementioned clustering-based sampling to obtain a set of training data. A machine learning model is trained on this data set. We obtain the model's classification accuracy on a held-out test dataset or using K-fold cross validation on the training data.  Comparing with the classification accuracy from the previous run (set the accuracy to be 0 for the first run), if the accuracy improves, we will do an incremental sampling on all the clusters, per user defined sampling size. If the change of classification accuracy is within a predetermined threshold or the total sampling size reaches an upper bound, the sampling process will be stopped. 

\subsubsection{Metadata-based Sensitivity Prediction}
\label{modeling}

We use the newly sampled dataset from Section~\ref{sampling} to train a binary classification machine learning model. Each file is represented as a feature vector, derived from the file metadata, consisting of features such as tokens in file names, file extensions, paths, and sizes. The output is the classification label ``sensitive'' or ``non-sensitive''. Once the machine learning model is built, we can then apply it to classify the remaining files on the volume based on the available metadata obtained from the infrastructure scan in Section~\ref{scan}.

\section{Experiment}\label{exp}
\begin{figure}[h]
\begin{center}
\includegraphics[width=3.4in]{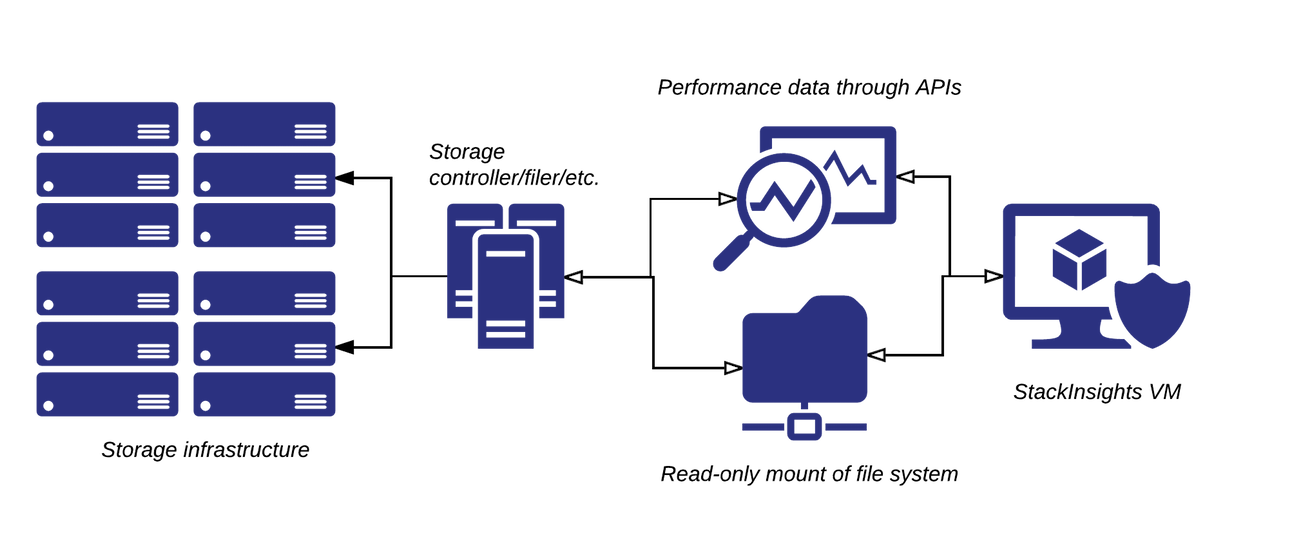}
\caption{Infrastructure, File Metadata and File Content Scan}
\label{fig:scan}
\end{center}
\end{figure}

Our environment consists of roughly 100 different accounts, with a wide variety of storage requirements. Each account has a mixture of file, block, and object storage.  For experiment, we choose a mid-size account, whose storage infrastructure is predominantly file storage. This account has two sites (data centers), each one with a set of NetApp filers in clustered mode (roughly 33.8 TB). We install one secure virtual machine inside each site preloaded with our StackInsights scanning codebase. We use those virtual machines to scan the environment and extract the necessary information for further analysis. Figure~\ref{fig:scan} shows a high-level diagram of our sample environment. The storage filer gives us a map of what the file storage infrastructure looks like. We first build a tree of the infrastructure by starting from the cluster and then down to the file level (i.e., $cluster~\Rightarrow~aggregate~\Rightarrow~volume~\Rightarrow~q-tree~\Rightarrow~file system~\Rightarrow~directory~\Rightarrow~file$). A collected file metadata example is shown in Table~\ref{filerecord}. In parallel, we poll the storage filers for IOPS for each storage volume. This will allow us to measure the IO density for each system, thereby, to determine how volatile a given volume or file share is.  

We use IBM's TADDM tool to get a picture of the different workloads running in the environment. Similarly, the system administrators also provide their valuable feedback, as they are  able to help us narrow our scanning scope. Because the filers in the environment are all NetApp filers, we use the ONTAP APIs to extract file metadata, as well as performance metrics from the filers. All machine learning algorithms are implemented based on the python scikit-learn library. 


\subsection{IOPS and file metadata collection}\label{iops_and_metadata}
\begin{figure}[h]
\begin{center}
\includegraphics[width=3.2in]{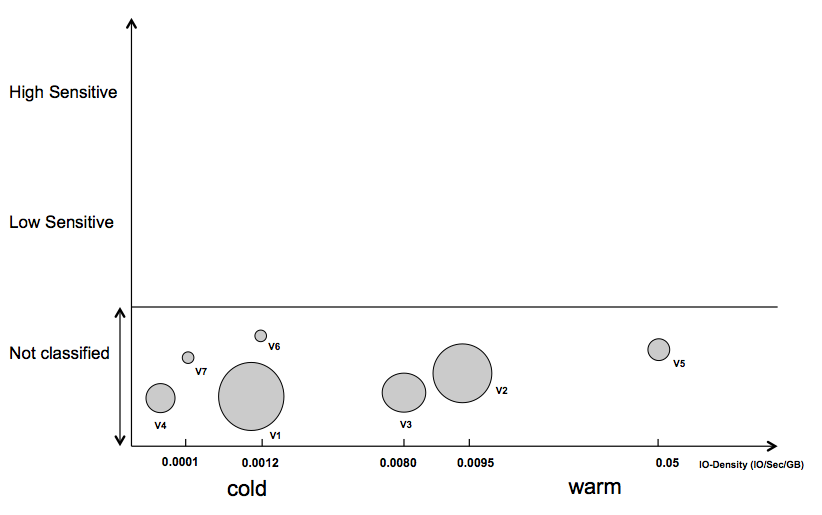}
\caption{Volumes are aligned according to IO density}
\label{fig:iops}
\end{center}
\end{figure}

The IOPS for each volume is collected over a four-week time window. We then compute the hourly-average IO density (IO per second per GB) for each volume. Table~\ref{iops} shows the total size of each volume as well as its corresponding IO density range. As we can see, except V5, the IO density of all the other volumes is between 0.0 and 0.01, which is relatively cold. Based on the storage performance metric, we may recommend the tier 3 storage type for V5 and the nearline or inactive storage type (e.g., archiving) for the other volumes. In Figure~\ref{fig:iops}, all the volumes are first aligned along the $X$-axis according to their IO density. Note that since the highest IO density is only around 0.05, we indicate the corresponding hotness as ``warm'' on the $X$-axis.

\begin{table}[htp!]
	{\small
	\caption{Volume IO density} 
	\label{iops}
	\begin{center}
	\begin{tabular}{| c | c | c |}
			\hline
			Volume name & Total size (TB) & I/O density range \\
			\hline
			V1 & 13.66 & 0 - 0.01 \\
			\hline
		        V2 & 12.32 & 0 - 0.01 \\
		        \hline
		        V3 & 6.06  & 0 - 0.01 \\
		        \hline
		        V4 & 1.14  & 0 - 0.01 \\
		        \hline
		        V5 & 0.66  & 0.01 - 0.1 \\
		        \hline
		        V6 & 0.01  & 0 - 0.01 \\
		        \hline
		        V7 & 0.01  & 0 - 0.01 \\
		        \hline
        \end{tabular}
        \end{center}
        }
\end{table}

\renewcommand{\arraystretch}{1.15}
\begin{table*}
	\caption{A file metadata example} 
	\label{filerecord}
	\resizebox{\textwidth}{!} {
		\begin{tabular}{| l | l | l | l | l | l | l | l | l | l | l |}
			\hline
			Attribute & File name & File extension & File path & Last accessed time & Creation time & Changed time & Last modified time & File size (bytes) & Bytes used \\
			\hline
			Value & Feedback Survey 2015.docx & .docx & /path/to/file/location & 2015-10-21 09:28:37.00 & 2015-10-21 09:28:31.00 & 2016-08-08 20:51:35.00 & 2015-10-21 09:28:37.00 & 20195	& 20480 \\
			\hline
		\end{tabular}
	}
\end{table*}

\renewcommand{\arraystretch}{1.15}
\begin{table*}
	\caption{A volume metadata example} 
	\label{volumerecord}
	\resizebox{\textwidth}{!} {
		\begin{tabular}{| l | l | l | l | l | l | l | l | l | l | l |}
			\hline
			Attribute & Volume size & Total file count & Total file size & Top3ExtensionbySize & Top3ExtensionbyCount& NotModifiedin1YearCount & NotModifiedin1YearSize & 	NotModifiedin3YearCount \\
			\hline
			Value &			6.06 TB & 3,627,061 & 2.64 TB & nsf, zip, xls & doc, xls, pdf & 94.18\% & 89.44 \% & 	0.0\% \\		
			\hline
			Attribute & NotModifiedin3YearSize &  NotAccessedin1YearCount & notAccessedin1YearSize & NotAccessedin3YearCount & NotAccessedin3YearSize & NotAccessedAfter2WeekCount & NotAccessedAfter2WeekSize & IO Density \\
			\hline
			Value &			0.0\% &  81.03\% & 76.28\% & 0.0\% & 0.0\% & 49.10\% & 53.13\% & 7.89E-3 \\
			\hline
		\end{tabular}
	}
\end{table*}

We leverage the NetApp ONTAP APIs to extract the metadata of all the files on these volumes. In total, we extract the metadata from more than 13 million files. One example is shown in Table~\ref{filerecord}, where ``Last accessed time'' is the time stamp that the file was last accessed, ``Creation time'' is the time stamp that the file was created, ``Changed time'' is the time stamp that the file metadata (e.g., file name) was changed, ``Last modified time'' is the time stamp that the file itself (not metadata) was last modified, ``File size'' represents the size that was allocated to the file on disk, and ``bytes used'' represents the bytes that were actually written in that file. The volume level metadata is then obtained by aggregating the metadata of all the files on the same volume. One volume metadata example is shown in Table~\ref{volumerecord}. The ``Top3ExtensionbySize'' attribute is the top three file extensions ranked by their total sizes. For instance, in Table~\ref{volumerecord}, ``.nsf'', ``.zip'', and ``.xls'' are the top three file extensions on that volume in terms of their total sizes. ``NotModifiedin1YearCount'' is the percentage of files on that volume that have not been modified in the past one year in terms of file counts. Similarly, ``notAccessedin1YearSize'' is the percentage of files on that volume that have not been accessed in the past one year in terms of file sizes. All the other attributes are likewise.  The volume metadata can provide us some further insights. For example, for one volume, we discover that about half number of the files have not been accessed in the past one year, but they take about 98\% storage capacity in size.

\subsection{Volumes clustering}
\begin{figure}[h]
\begin{center}
\includegraphics[width=3.2in]{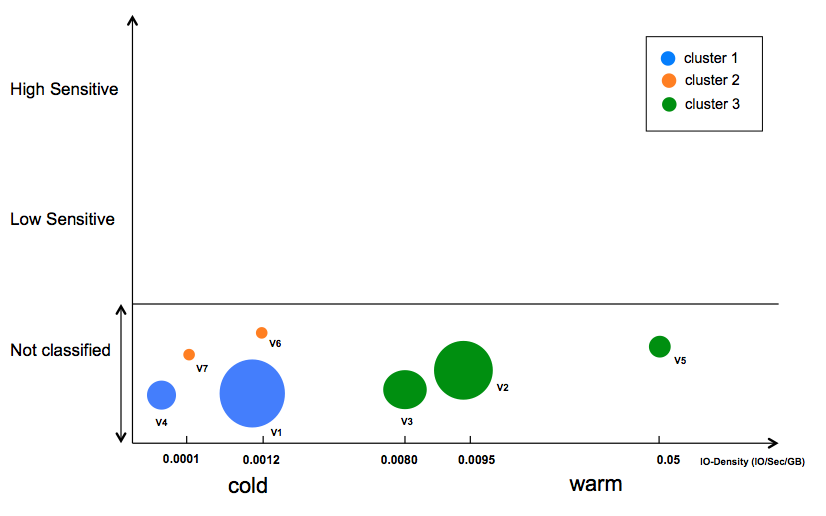}
\caption{Volume clustering}
\label{fig:vcluster}
\end{center}
\end{figure}

After the volume level metadata is obtained, we apply K-means to do clustering on all the volumes. Due to the relatively small number of volumes, we empirically set $K=3$. The optimal value of $K$ can be determined by the elbow method, which considers the percentage of variance explained by the clusters against the total number of clusters. The first clusters will explain much of the variance. The optimal number of clusters is chosen at the point where the marginal gain will drop. Figure~\ref{fig:vcluster} shows the clustering results when $K=3$. Note that we have not considered the data sensitivity yet, so all the volumes are still aligned along the $X$-axis. 

We select a representative volume from each cluster, which is defined as the one with the minimum total distance to the other volumes in the same cluster. A sensitivity analysis is performed on each representative volume. The volumes in the same cluster are assumed to share similar sensitivity scores.

\subsection{Metadata-based Sensitivity Prediction}
We present in this section the experimental results of the metadata-based sensitivity prediction. 

\subsubsection{Training Data}
For each representative volume, we build a machine learning model to learn the sensitivity of all the files. One of the selected volume from the first data center contains 3.9 million files, which are 2.64 TB in size. Through the infrastructure scan introduced in Section~\ref{scan}, we obtain the metadata of all the files. We randomly select a subset of the files to crawl their content and determine the sensitivity using the dictionary based approach introduced in Section~\ref{datamap}. Specifically, we use Apache Tika to scan the file content. A file is considered as sensitive if it contains any sensitive information listed in Table~\ref{tb:sendic}. We finally obtain a set of 114,854 files with both their metadata and sensitivity labels, out of which, 66,221 (57.65\%) files labelled as sensitive and the rest as non-sensitive. Similarly, we obtain another set of 39,571 files with both their metadata and sensitivity labels on a representative volume from the second data center. This data set includes 21,284 sensitive files (53.79\%). In the following, we will refer to these two data sets as dataset I and dataset II. Unless noted, the percentage of sensitive files in these two datasets remain as 57.65\% and 53.79\%, respectively. 

\subsubsection{Feature Engineering}
Given the training data, we derive features from file metadata for the classification model. Specifically, the features are divided into several categories: file name, file extension, file path, file size related, and time related. We will briefly introduce each feature category.

\textbf{Name Features:} \hspace{1pt} We extract the name of the file in plain text. The file name can contain textual information that indicate the file sensitivity. For example, a file named ``patent disclosure review\_Feb2\_2015.docx'' probably contains intellectual property information and should be considered as sensitive. In order to exploit those textual information, we model each file name using the bag-of-word approach and represent it as a vector $v = [v_1, ..., v_n]$, where $n$ is the size of the vocabulary and $v_i$ is the frequency of word $i$ in the file name. Before computing the feature vectors, we clean the file names by removing all numeric characters, punctuation marks, and stop words. Finally, we obtain the textual feature vector, whose vocabulary size is around 28,000 for dataset I and 15,000 for dataset II.

\textbf{Path Features:} \hspace{1pt} The file paths are extracted from the file system, indicating where the files are stored. In order to exploit this feature, we transform the file path into a binary vector. We choose a parameter $d$ which controls the depth of the folders we explore. We extract all the folders that are $d$ levels away from the root. Assuming that we have found $m$ folders, we store them in a list $l=[l_1,...,l_m]$. Then, for a given file $f$, we represent it as a feature vector $v=[v_1,...v_m]$ of length $m$ where $v_i = 1$ if $f$ belongs to folder $l_i$, otherwise 0.

 
\textbf{Extension Features:} \hspace{1pt} We expect that certain file extensions are more likely to be sensitive than others. In order to use this feature, we apply a similar procedure as for the file paths. The extension features are encoded as a binary vector. We collect all the extensions that belong to our training set and store them in a list $e=[e_1,...,e_m]$ where m is the number of extensions we have collected. For a given file $f$, we will represent it as a feature vector $v = [v_1, ..., v_m]$, where $v_i = 1$ if $f$ has extension $e_i$, otherwise 0.

\begin{table}[tp]
\caption{Feature summary}
{\small
\begin{center}
\label{tb:feature_size}
  \begin{tabular}{| c | c | c|}
  \hline
    Feature category & dataset I & dataset II \\ \hline
    File name & 29736  & 14861\\ \hline
    File path & 788  & 315\\ \hline
    File extensions & 1170 &  316 \\ \hline
    File size related & 2 & 2 \\ \hline
    Time related & 3 & 3 \\ \hline
    Total feature size & 31699 & 15497 \\ \hline
  \end{tabular}
\end{center}
}
\end{table}

\textbf{Size Related Features:} \hspace{1pt}  We have access to two features: the size of the file and the bytes used allocated to the file. The file size represents the size that was allocated on disk to the file, while the bytes used represents the bytes that were actually written. 

\textbf{Time Related Features:} \hspace{1pt}  We include three time related features for each file: specifically, the time difference between the last accessed time and the creation time, the difference between the changed time and the creation time, and the difference between the last modified time and the creation time. All the differences are in the number of days. Please refer to section~\ref{iops_and_metadata} for the detailed meaning of these time stamps. 

\textbf{Feature Summary:} \hspace{1pt} After the features in each category are collected, we concatenate them into a larger feature vector to represent each file. Because the size of the file path feature grows exponentially with the depth $d$, we choose $d=2$ empirically in our experiments. All the features are normalized into the range $[0,1]$. Table~\ref{tb:feature_size} summarizes the total feature size for each category and the overall size of the feature vector.

\subsubsection{Feature Selection}
We use feature selection to get a more in-depth view of all the features that are significant in the machine learning model. In particular, we investigate which features and what types of features are the most significant. We apply mutual information~\cite{mi2005} to select the top features. Table~\ref{tb:mi} shows the top 10 selected features.

\begin{table}[tp]
\caption{Top ten features}
{\small
\begin{center}
\label{tb:mi}
  \begin{tabular}{| c | c | c |}
  \hline
    Feature Type & dataset I & dataset II \\ \hline
	Extension &  .url  & .xls \\ \hline
	Extension &  .properties & .txt \\ \hline
	Extension &  .mdm & .html \\ \hline
	Extension &  .pas & .net \\ \hline
	File Size & - & - \\ \hline
	Bytes Used & - & - \\ \hline
	Last access time diff & - & - \\ \hline	
	Change time diff & - & - \\ \hline	
	Last modified time diff & - & -  \\ \hline
	Text & ``feature" & ``username" \\ \hline
  \end{tabular}
\end{center}
}
\end{table}
Among the top 10 features, file extension features take the most percentage. In addition, the two file size related and three time related features are also significant. Text tokens ``feature'' and ``username'' are also among the top 10. Note that we use ``username'' to replace an actual username for privacy concerns. We do not find any file path features in the list, which may indicate that the location of the file in the filesystem carries less significance than it's size, time, name, or extension in the prediction. For example, one particular folder may include both sensitive and non-sensitive files. 

Table~\ref{feature_category} shows the number of selected features in each category when we vary the total number of top selected features. Again we notice that file extensions and text tokens in file names are significant features, while the file paths do not appear in the top list.

\begin{table}[htp!]
	{\small
	\caption{Top feature categories} 
	\label{feature_category}
	\begin{center}
	\begin{tabular}{| c | c | c |}
	\hline
			Feature Type & Top 100  & Top 500 \\ \hline
			Extension & 27  & 79\\ \hline
			File size related & 2 & 2 \\ \hline
			Time related  & 3 & 3\\ \hline
			Text & 68 & 416 \\ \hline
        \end{tabular}
        \end{center}
        }
\end{table}

\subsubsection{Prediction Models}
After all the features are extracted, we build machine learning models on our training data and apply them to predict the file sensitivity based on meta data. Specifically, we compare the performance of several well-known classification models: Naive Bayes, Logistic Regression, Support Vector Machines (SVM), and Random Forest.

All the experiments are conducted using the 10 fold cross validation. Since only file size related and time related features have numerical values (in total five), among the large number of features, we apply multinomial Naive Bayes, the feature distributions of which are modeled as multinomial distributions rather than the gaussian distributions. Naive Bayes has the advantage of having no parameters to optimize on. Logistic Regression has only one parameter that we need to tune: the regularization parameter $C$. In order to select the optimal value $C$, we run grid search using 10-fold cross-validation on multiple values of $C$. We find that the best $C$ value is $0.9$. For SVM, the linear kernel is selected. In practice, we find that the RBF kernel takes a very long time to converge. The optimal regularization parameter $C$ is selected following the same procedure as with Logistic Regression. The optimal $C$ value is set to be 0.8 for linear SVM. We use the default parameter setting for Random Forest, where the number of tree in the forest is 10, no maximum tree depth constraint, and the samples are drawn with replacement. Table~\ref{performance} shows the performance of each model in terms of overall accuracy, precision, recall, and F1 score. In our classification problem, the positive class is ``sensitive'' while the negative class is ``non-sensitive''. Specifically, the precision is defined as the ratio $tp / (tp + fp)$, where $tp$ is the number of true positives and $fp$ the number of false positives. The recall is the ratio $tp / (tp + fn)$, where tp is the number of true positives and fn the number of false negatives. Accuracy is defined as the ratio between the number of samples that are correctly classified and the total number of samples. The F1 score is computed as $2 \times (precision \times recall) / (precision + recall) $,  which is a weighted average of the precision and recall.

In Table~\ref{performance}, the percentages of sensitive files in dataset I and II are 57.65\% and 53.79\%, respectively. Therefore, the classes in the training data are roughly balanced. As we can see, Random Forest has the best performance among all the models over all the metrics. The precision and recall on dataset I are above 90\%. In contrast to other models, Random Forest, as an ensemble method, combines the predictions of several based estimators, i.e., decision trees. Each tree in the ensemble is built from a sample drawn with replacement from the training set. When splitting a node during the construction of the tree, the split is chosen as the best split among a random subset of the features. Since both the feature size and sample size are large in our classification, as a result of this randomness, the variance of the forest is reduced due to averaging, hence yielding an overall better model. 

In practice, the percentage of sensitive files in the training data depends on the specific domain and sensitivity labeling. In some domain, the sensitivity labelling may be stringent, resulting in a relatively small percentage of sensitive files. We also design experiments to test the performance of machine learning models for such case. Specifically, we only use the phone regular expression in Table~\ref{tb:sendic} to do the labelling, discarding the other sensitive information, which yields 25,256 (21.99\%) sensitive files for dataset I. We apply four different machine learning models and report the results in Table~\ref{performance2} for imbalanced classes. The ``balanced'' classification mode is used for Logistic Regression, SVM, and Random Forest, where the values of prediction target $y$ are used to automatically adjust weights inversely proportional to class frequencies in the training data. As shown in Table~\ref{performance2}, Random Forest has the best performance among all the models over all the metrics, except recall. Due to space limit, we only show the results on dataset I. 

\begin{table}[htp!]
	{\scriptsize
	\caption{Models on datasets of balanced classes} 
	\label{performance}
	\begin{center}
	\begin{tabular}{| c | c | c | c | c |}
	\hline
			 Model (dataset I) &  Accuracy &  Precision & Recall &  $F1$ \\ \hline
			 Naive Bayes & 0.8044 & 0.8348 & 0.8238 & 0.8293 \\ \hline
			 Logistic Regression & 0.8115 & 0.8664 & 0.7959 & 0.8296 \\ \hline
			 SVM & 0.8309 & 0.8730 & 0.8269 & 0.8493 \\ \hline
		 	 Random Forest & \bf{0.9014} & \bf{0.9250} & \bf{0.9022} & \bf{0.9135} \\ \hline
        \end{tabular}	
        \vspace{7pt}
        
	\begin{tabular}{| c | c | c | c | c |}
	\hline
			 Model (dataset II) &  Accuracy &  Precision & Recall &  $F1$ \\ \hline
			 Naive Bayes & 0.7780 & 0.7855 & 0.8081 & 0.7966 \\ \hline
			 Logistic Regression & 0.7923 & 0.8350 & 0.7652 & 0.7985 \\ \hline
			 SVM & 0.8055 & 0.8493 & 0.7762 & 0.8111 \\ \hline
		 	 Random Forest & \bf{0.8739} & \bf{0.8926} & \bf{0.8703} & \bf{0.8813} \\ \hline
        \end{tabular}
        \end{center}
        }
\end{table}

\begin{table}[htp!]
	{\scriptsize
	\caption{Models on dataset of imbalanced classes} 
	\label{performance2}
	\begin{center}
	\begin{tabular}{| c | c | c | c | c |}
	\hline
			 Model (dataset I) &  Accuracy &  Precision & Recall &  $F1$ \\ \hline
			 Naive Bayes & 0.8813 & 0.8122 & 0.5988 & 0.6894 \\ \hline
			 Logistic Regression & 0.8538 & 0.6270 & \bf{0.8274} & 0.7134 \\ \hline
			 SVM & 0.8774 & 0.6836 & 0.8237 & 0.7471 \\ \hline
		 	 Random Forest & \bf{0.9361} & \bf{0.9176} & 0.7795 & \bf{0.8429} \\ \hline
        \end{tabular}
        \end{center}
        }
\end{table}

As we can see, Random Forest has very robust performance, and consistently outperforms the other methods in both balanced and imbalanced classifications. Previously, we have used 10-fold cross validation to test the prediction performance, which uses 90\% data for training and the remaining for testing. We now vary the percentage of training and testing, and check how Random Forest performs if relatively small percentage of data is used for training. Table~\ref{performance3} shows the performance of Random Forest with two-fold cross validation on both dataset I and II, where 50\% data are used for training and 50\% for testing. Again, Random Forest has shown good performance on all the metrics, even better than the other models when 90\%  data are used for training. 

\begin{table}[htp!]
	{\scriptsize
	\caption{Performance with two-fold cross validation} 
	\label{performance3}
	\begin{center}
	\begin{tabular}{| c | c | c | c | c |}
	\hline
			 &  Accuracy &  Precision & Recall &  $F1$ \\ \hline
			Random Forest (dataset I) & 0.8790 & 0.9072 & 0.8802 & 0.8935 \\ \hline
			Random Forest (dataset II) & 0.8583 & 0.8820 & 0.8503 & 0.8658 \\ \hline
        \end{tabular}
        \end{center}
        }
\end{table}

To have a detailed analysis of the classification results, we show the confusion matrices of Random Forest (one fold in a two-fold cross validation) in Table~\ref{nb_cm}, which allows us to see how well the model performs on the classification of each class. Overall, the error ratios on false positives and false negatives are balanced on both datasets. 

\begin{table}[htp!]
	{\small
	\caption{Model confusion matrices} 
	\label{nb_cm}
	\begin{center}
	\begin{tabular}{| c | c | c |}
				\hline
			Random Forest (dataset I) & Non-Sensitive &  Sensitive \\ \hline
    			Non-Sensitive & 21493 (0.88) & 2824 (0.12)  \\ \hline
    			Sensitive & 3999 (0.12) & 29112 (0.88) \\ \hline
        \end{tabular}           
        \vspace{7pt}
        
        	\begin{tabular}{| c | c | c |}
				\hline
			Random Forest (dataset II) & Non-Sensitive &  Sensitive \\ \hline
    			Non-Sensitive & 7897 (0.86) & 1247 (0.14)  \\ \hline
    			Sensitive &1535 (0.14) & 9107 (0.86) \\ \hline
        \end{tabular}      

	\end{center}

        }
\end{table}

\textbf{Prediction Model Usage}: Note that we do not intend to use the above prediction model to completely replace the traditional content scanning method, such as dictionary based method. As we can see, the prediction model is based on meta data and cannot achieve 100\% accuracy. In data governance and security, the misclassification of sensitive data can be catastrophic for an organization. For example, critical IP documents leak or data compliance violation can lead to serious legal consequences. Therefore, a thorough sensitivity screening can be performed to make sure all the sensitive information are identified. From the machine learning model, all the files that have been predicted as sensitive will be labelled as sensitive data. We can then perform intensive content scanning method on all the files that are predicted to be non-sensitive. For example, after applying Random Forest on dataset I, 25,492 files are predicted as non-sensitive. The content scanning based method will then be applied to these files, so that the 3,999 mis-classified sensitive files can be identified. In contrast to the content scanning of all the 57,428 files, we now only need to perform content scanning on 25,492 files (44.38\%), significantly smaller than the original number of files. There are certainly non-sensitive files mis-calssified as sensitive files. For example, 2824 non-sensitive files are mis-classified as sensitive files. As a result, they are ``over-protected''. However, the percentage of such files only takes 4.91\% of the total files. As a simple comparison baseline, with the percentage of sensitive files in the training data (i.e., 57.65\%) for dataset I, a user can randomly selects 57.65\% data, and label them as sensitive and the remaining as non-sensitive, without using the prediction model. They can then perform content scanning on the previously labelled non-sensitive files in order to identify any sensitive information. Note that in this baseline, among the 57.65\% files that are labelled as sensitive, 42.35\% files are actually non-sensitive (based on the percentage of non-sensitive files in the training data), therefore, $57.65\% \times 42.35\% = 24.41\%$ amount of non-sensitive files are misclassified as sensitive, therefore ``over-protected'', in contrast to 4.91\% that are misclassified by the prediction model. 

\subsubsection{Prediction Ranking and Running Time}
After the machine learning model is trained, we apply it to predict the sensitivity of the remaining files on the same volume. For dataset I, we apply Random Forest to predict the sensitivity of the remaining 3.9 million files and measure its running time. On a local machine with 2.5 GHz Intel Core i7 CPU and 16GB of RAM, the total running time is 112 minutes. The model classifies 1,804,798 (46.33\%) files as sensitive and 2,089,913 (53.67\%) files as non-sensitive. As a comparison, the content scanning based approach took more than 30 hours to process 228,000 files, which is only 5.85\% of the 3.9 million files. StackInsights reduces the total running time by orders of magnitude. 

We also apply the trained learning model to predict the sensitivity of files on other volumes in the same data center. Table~\ref{summary} shows the prediction results: the predicted sensitive files \#, volume sensitivity, and running time in seconds. As we can see, V1 and V4 have sensitivity close to 1.00. They also belong to the same cluster in Figure~\ref{fig:vcluster}. V2, V3, and V5 have sensitivity between 0.45 and 0.70, which are in the same cluster. V6 and V7 are in the same cluster, with predicted sensitivity 0.5758 and 0.1728, respectively. 

\begin{table}[htp!]
	{\scriptsize
	\caption{Prediction results on all the volumes} 
	\label{summary}
	\begin{center}
	\begin{tabular}{| c | c | c | c | c | c |}
	\hline
			 &  Total file \# &  Sensitive file \# & Sensitivity & Running time (sec) \\ \hline
			 V1 & 400415 & 385369 &  0.9624 & 286.28 \\ \hline
			 V2 & 7798224 & 5501763 &  0.7055 & 5993.20 \\ \hline
			 V3 &  3894711 & 1804798 &  0.4633 & 6720 \\ \hline
			 V4 & 170808 & 170808 &  1.0000 & 134.01 \\ \hline
			 V5 & 1481322 & 902804 &  0.6095 & 1133.58 \\ \hline
			 V6 & 686 & 395 & 0.5758 & 0.51 \\ \hline
			 V7 & 81 & 14 & 0.1728 & 0.06 \\ \hline
        \end{tabular}
        \end{center}
        }
\end{table}

\subsubsection{Migration Insights}
After the sensitivity of all the files on a storage volume is predicted, we can compute sensitivity scores at different levels. The sensitivity score is defined as the number of sensitive files divided by the total number of files at a targeted level. We give two examples: volume level and user level. Similarly, we can also obtain data hotness (e.g., storage performance metrics) at different levels. From the Infrastructure scan, we compute the IO density for volumes. For the user level, we use the metric, the percentage of files that have not been accessed in past one year, to represent the data hotness for a particular user folder. By correlating data sensitivity and hotness, StackInsights can provide hybrid could migration insights. In addition, StackInsights can be integrated with data movement tools, such as DataDynamics~\cite{datadynamics}, in order to automate the entire migration process from recommendation to action. 

\begin{figure}[htp!]
\begin{center}
\includegraphics[width=3.2in]{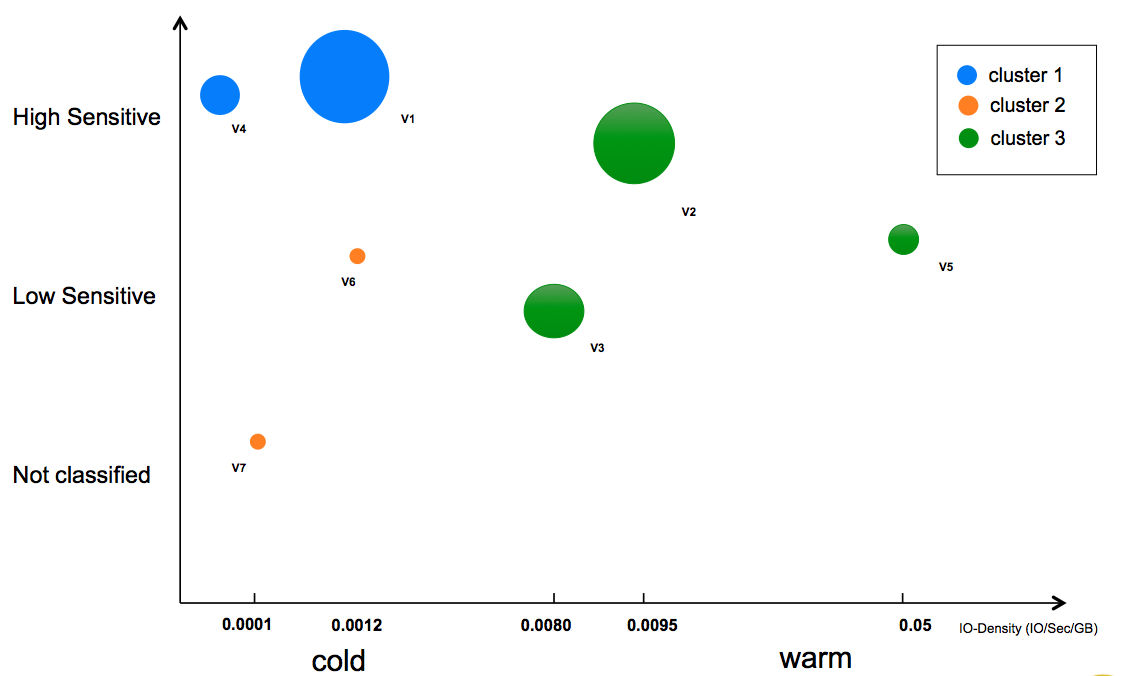}
\caption{Volume sensitivity and IO density map}
\label{fig:vpredict}
\end{center}

\end{figure}

\textbf{Volume Sensitivity and Hotness Map:} \hspace{1pt} We get the sensitivity score for each volume from the prediction results. As shown in Figure~\ref{fig:vpredict}, all the volumes are ordered based on their sensitivity level and hotness. Therefore, all the volumes which are cold and low  sensitive can be migrated to the public cloud. The remaining should be migrated to the private cloud or remain on premise.

\begin{figure}[htp!]
\begin{center}
\includegraphics[width=2.7in]{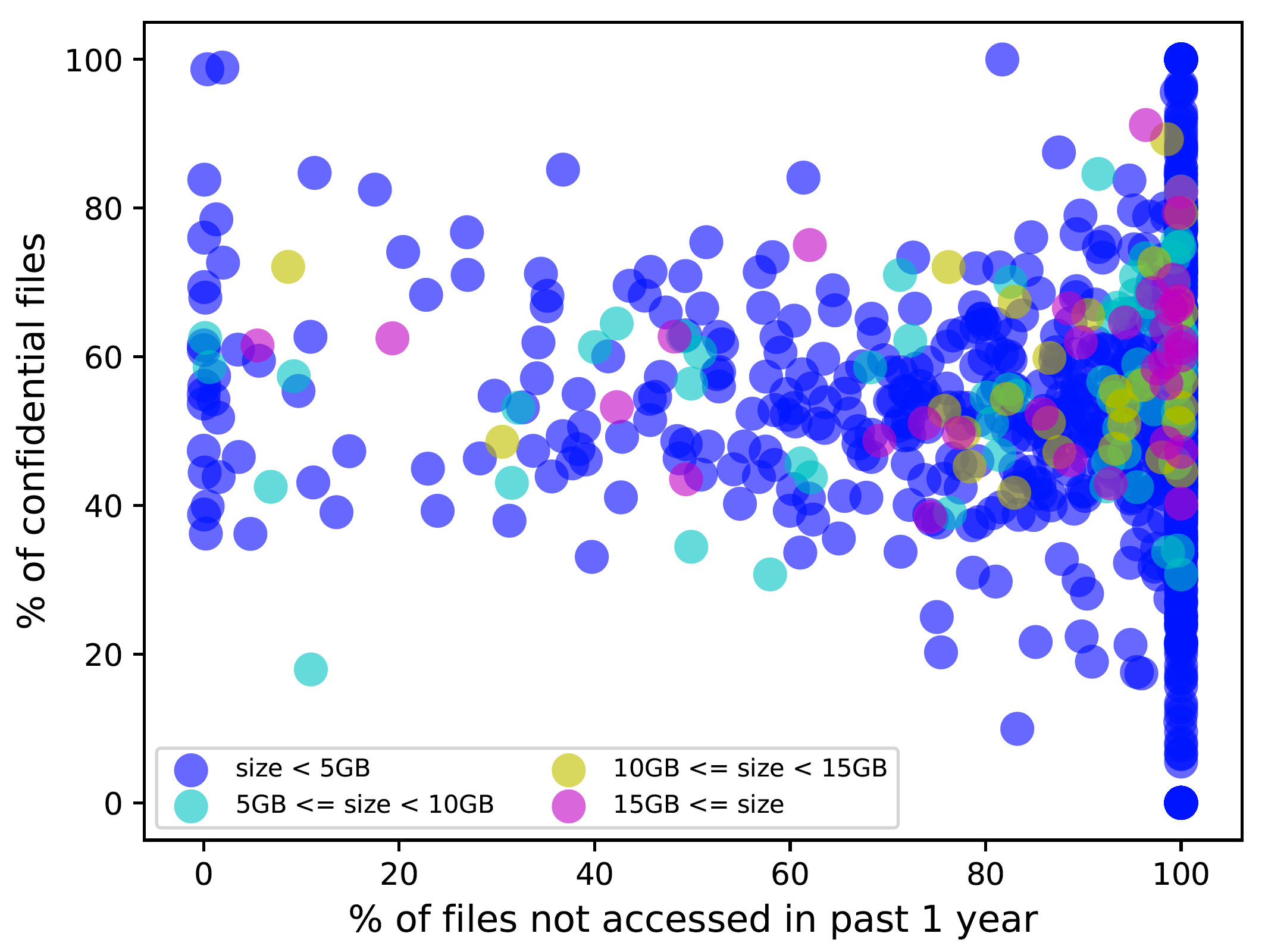}
\caption{User sensitivity and data hotness map}
\label{fig:usermap}
\end{center}
\end{figure}

\textbf{User Sensitivity and Hotness Map:} \hspace{1pt} Similar as the volume level analysis, we can also obtain the data sensitivity and hotness map on the user level. The data hotness for a particular user folder is computed as the percentage of all the associated files that have not been accessed in the past one year. Specifically, we find 1,060 user folders on volume V3. We compute the sensitive score for each user folder based on the predicted file sensitivity, as well as the data hotness using the file metadata collected from infrastructure scan. The user sensitivity and hotness map is shown in Figure~\ref{fig:usermap}. As we can see, there are many user folders that have not been accessed in the past one year. The percentage of sensitive files under these folders vary though. For user folders at the bottom right (cold and low sensitive), they may be eligible to be migrated to the public cloud. For those at the top left (hot and high sensitive), they should be migrated to the private cloud or remain on premise.

\section{Conclusions and Future Work}
\label{conclusion}

We have introduced StackInsights, a cognitive learning system which automatically analyzes and predicts the cloud readiness of workloads. StackInsights correlates the metrics from application, data, and infrastructure layers to identify the business sensitivity of data as well as their hotness in terms of infrastructure performance, and provides insights into hybrid cloud migration. Given the scale of data and infrastructure, a machine learning model is developed in StackInsights to predict file sensitivity based on the metadata. In contrast to traditional approach which requires intrusive and expensive content scanning, StackInsights significantly reduces the total running time for sensitivity classification by orders of magnitude, therefore, is scalable to be deployed in large scale IT environment. As more and more enterprises are committing to hybrid cloud architecture, StackInsights can help accelerate this digital transformation in their organizations. 

Our current system is mainly focused on understanding the sensitivity of textual files. There are many different types of data that can contain sensitive information, such as images, videos, and audios. In the future, we can leverage IBM Watson services~\cite{watson} to analyze the sensitive content from these multimedia data. Similarly, we can predict their sensitivity based on the meta level information. Last but not the least, the cognitive learning capabilities of StackInsights can be greatly enhanced by collecting more metadata across the stack. The more metadata we can collect, the more accurate the prediction model will be.



%

\bibliographystyle{IEEEtran}
\bibliography{IEEEabrv,stack_insight_paper}

\end{document}